\documentclass[sigconf, nonacm]{acmart}

\setcopyright{none}
\renewcommand\footnotetextcopyrightpermission[1]{}
\usepackage{multirow}
\usepackage{enumitem}

\AtBeginDocument{%
  }

\definecolor{amber}{rgb}{1.0, 0.49, 0.0}
\definecolor{dodgerblue}{RGB}{30, 144, 255}
\definecolor{violet}{RGB}{238,130,238}
\definecolor{my_green}{RGB}{113,173,71}
\definecolor{my_blue}{RGB}{44,115,182}

\newcommand{\reffig}[1]{\textcolor{black}{Fig.~\ref{fig:#1}}} 
\newcommand{\refsec}[1]{\textcolor{black}{Sec.~\ref{sec:#1}}}
\newcommand{\reftab}[1]{\textcolor{black}{Tab.~\ref{tab:#1}}}

\newcommand{\eg}[1]{\textcolor{black}{\textit{e.g.,~}}}
\newcommand{\ie}[1]{\textcolor{black}{\textit{i.e.,~}}}

\begin{document}

\title{Text-Guided Vector Graphics Customization}


\author{Peiying Zhang}
\affiliation{
 \institution{City University of Hong Kong}
  \city{Hong Kong}
  \country{China}
 }
\email{zhangpeiying17@gmail.com}

\author{Nanxuan Zhao}
\affiliation{
 \institution{Adobe Research}
  \city{San Jose}
  \country{USA}
 }
\email{nanxuanzhao@gmail.com}

\author{Jing Liao}
\authornote{Corresponding author}
\affiliation{
 \institution{City University of Hong Kong}
 \city{Hong Kong}
 \country{China}
 }
\email{jingliao@cityu.edu.hk}



\begin{abstract}
Vector graphics are widely used in digital art and valued by designers for their scalability and layer-wise topological properties. However, the creation and editing of vector graphics necessitate creativity and design expertise, leading to a time-consuming process. In this paper, we propose a novel pipeline that generates high-quality customized vector graphics based on textual prompts while preserving the properties and layer-wise information of a given exemplar SVG.
Our method harnesses the capabilities of large pre-trained text-to-image models. By fine-tuning the cross-attention layers of the model, we generate customized raster images guided by textual prompts. To initialize the SVG, we introduce a semantic-based path alignment method that preserves and transforms crucial paths from the exemplar SVG. Additionally, we optimize path parameters using both image-level and vector-level losses, ensuring smooth shape deformation while aligning with the customized raster image.
We extensively evaluate our method using multiple metrics from vector-level, image-level, and text-level perspectives. The evaluation results demonstrate the effectiveness of our pipeline in generating diverse customizations of vector graphics with exceptional quality. 
The project page is \textcolor{blue}{ \url{https://intchous.github.io/SVGCustomization}}.
\end{abstract}

\ccsdesc[500]{Computing methodologies~Graphics file formats}
\ccsdesc[500]{Computing methodologies~Image processing}
\ccsdesc[300]{Applied computing~Arts and humanities}

\keywords{Vector Graphics, SVG, Diffusion Model, Image Vectorization}

\begin{teaserfigure}
  \includegraphics[width=\textwidth]{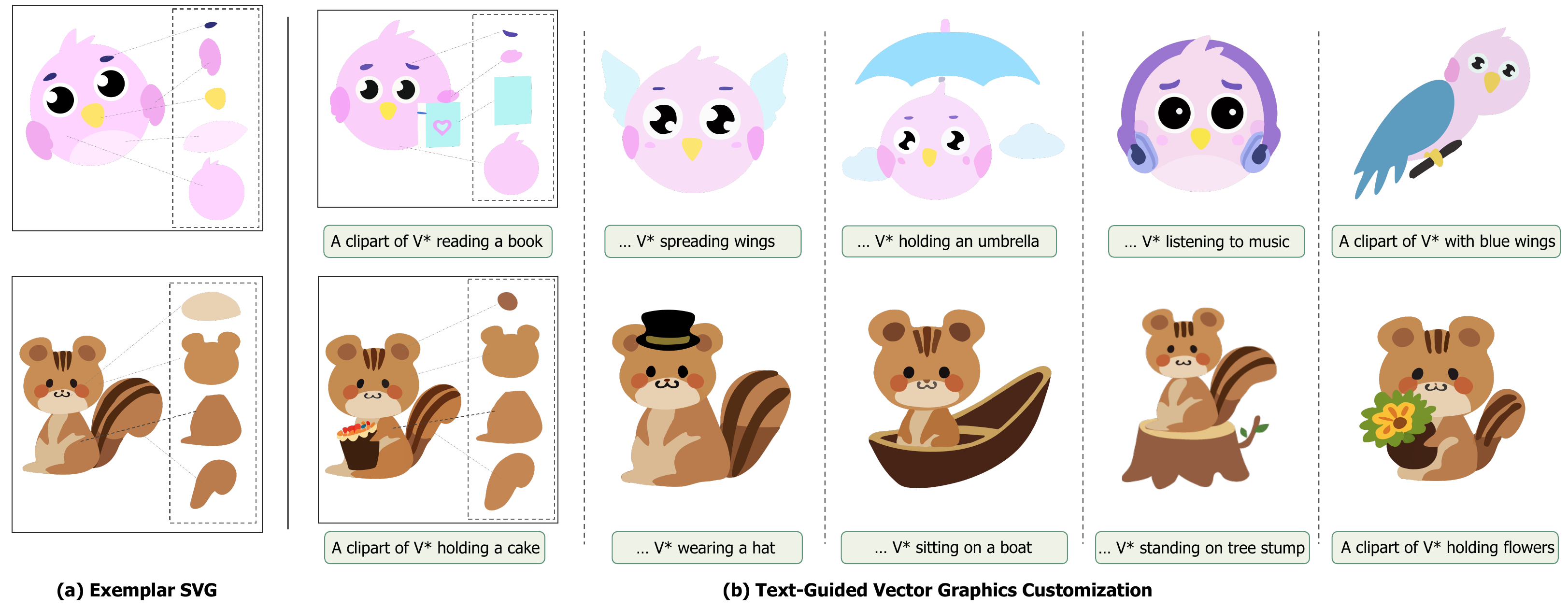}
  \caption{Text-guided vector graphics customization results. Given an exemplar vector graphic in SVG format (a), our model can generate customized vector graphics (b) based on diverse text inputs while keep the visual identity ($V^{*}$) of the exemplar SVG. Exemplar SVGs: the $1^{st}$ row is from \copyright{iconfont}; the $2^{nd}$ row is from illustac creator \copyright{Zarame}. }
  \label{fig:teaser}
\end{teaserfigure}

\maketitle

\section{Introduction}
\label{sec:introduction}
Vector graphics is a type of art composed of geometric shapes, such as lines, curves, and polygons. It plays essential roles under various practical scenarios (\eg~animations, cartoons, and graphic designs) due to its scalability, precision, and versatility for creating high-quality visuals and designs~\cite{quint2003scalable}. With the feature of being resolution-independent, vector graphics ensures consistent image quality across different media and devices, and it also enables easy manipulation, making it efficient to create intricate designs. However, creating high-quality vector graphics requires expertise and can be time-consuming~\cite{diebel2008bayesian}. Inspired by the recent success of text-to-image (T2I) generation \cite{ramesh2021zero, rombach2022high,  ruiz2022dreambooth}, one question raises whether the model can assist in automating the creation of vector graphics.

One approach for adapting from text-to-image to text-to-vector (T2V) is to first generate raster images from text and then convert them into vectors using existing image vectorization methods such as Potrace~\cite{selinger2003potrace}, Effective Vectorization~\cite{yang2015effective}, Photo2Clipart~\cite{favreau2017photo2clipart}, or Layer-Wise Vectorization~\cite{ma2022towards}. However, image vectorization methods often generate undesired vector elements (e.g., irregular shapes or overcomplicated paths) and layer relationships, complicating further graphic manipulation. Recently, a new category of T2V methods (e.g., VectorFusion \cite{jain2022vectorfusion}) has emerged that directly optimizes the paths of vector graphics using some pre-trained vision-language models such as the CLIP model\cite{radford2021learning} or the Diffusion model\cite{rombach2022high}. These methods can generate diverse vector graphics given the same text input. However, restricting the appearance of generated subjects and ensuring path regularity can be challenging with these methods.

To address the aforementioned issues in T2V, we propose a novel task called text-guided vector graphics customization. Following the previous works \cite{ma2022towards, frans2022clipdraw}, we use Scalable Vector Graphics (SVGs) with a set of parametric shape primitives (\eg~cubic Bézier curves) to represent vector graphics. Our objective is to generate customized SVGs that maintain the visual identity, vector properties, and layer relationships of an exemplar SVG, based on a given text prompt. Such a method can benefit a lot of practical use cases, imagining that you have created a new character or logo (\eg~a cute bird), but want to adapt it into a different context (\eg~the cute bird reading a book) or generate more variations, as shown in \reffig{teaser}. However, this task poses several unique challenges. First, the visual identity of the input SVG should be well preserved in the customized results, which is a difficult task given only a single SVG input. Second, to fit into diverse contexts, the geometry and appearance features of the customized SVG can be really different, how to maintain the correct path relationships and incorporate newly added contents/paths at the same time remains unsolved. Lastly, the method must generate visually pleasing results alongside valid vector paths, further increasing the complexity of the task.

To this end, we propose a pipeline for text-guided vector graphics customization. We leverage a powerful pre-trained diffusion model ~\cite{rombach2022high} that has been proven to be efficient in multiple T2I generation and editing tasks. By fine-tuning the cross-attention layers of the pre-trained diffusion model with exemplar SVG and input text prompt, we first generate a customized raster image that preserves the visual identity of the exemplar. Inspired by experts' design process that reuses essential semantic elements across a series of SVGs to achieve coherence with the concept\cite{ambrose2019fundamentals}, we propose a semantic-based path alignment method that keeps important paths based on semantic correspondence between the exemplar and the customized raster image. We apply rigid transformations to these kept paths to provide a good initialization of the output customized SVG. Path parameters (e.g., control point positions and fill colors) of the SVG are further optimized with a set of image-level loss and vector-level loss to reconstruct the customized raster image while preserving path regularity. 
The image-level loss measures the distance between the rendered SVG and customized image in the CLIP latent space, promoting fidelity to the customized image. The vector-level loss incorporates a local Procrustes loss, enforcing path regularity by minimizing local deformation of control points.
After optimization, a customized SVG corresponding to the text prompt and preserving the visual identity of the exemplar SVG is generated.

We assess our method using vector-level, image-level, and text-level metrics, which confirm its effectiveness in generating high-quality text-guided vector graphics customizations. Our key contributions include:
\begin{itemize}[leftmargin=*]
  \item We propose the first pipeline for vector graphics customization that generates customized SVGs based on text prompts while preserving the visual identity of the exemplar SVG.
   \item We present a semantic-based path alignment method that allows for the reuse of paths in the exemplar SVG, thus better maintaining the valid path properties and layer relationships of the exemplar SVG in customized SVGs.    
  \item We propose a path optimization method for image vectorization that uses both image-level loss and vector-level loss to reconstruct the target image while retaining the path regularity.
\end{itemize}

\begin{figure*}[tbp]
  \centering
  \includegraphics[width=1.0\linewidth]{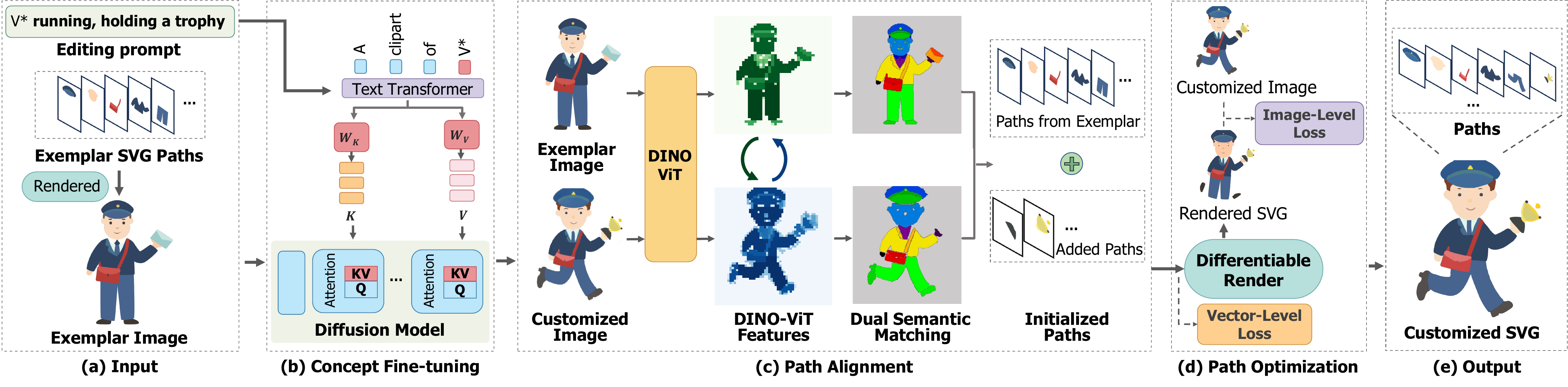}
  \caption{\label{fig:pipeline} 
  Text-Guided vector graphics customization pipeline.
  Given an exemplar SVG and a text prompt as input (a), the method consists of three stages. (b) Concept Fine-tuning: By fine-tuning a pre-trained T2I model, a customized raster image is generated based on the text prompt. 
  (c) Path Alignment: Important paths from the exemplar SVG are adapted based on semantic correspondences with the customized raster image, providing an initial customized SVG.
  (d) Path Optimization: The path parameters are optimized using both image-level and vector-level losses to produce the final customized SVG (e).
  The exemplar SVG is from envatoelements creator \copyright{Telllu}.
  }
\end{figure*}

\section{Related Work}
\label{sec:related_work}

\textbf{Text-to-Image Generation and Manipulation.}
Rapid progress of deep learning has driven advancements in T2I tasks. Early methods, such as text-conditional GANs~\cite{reed2016generative, li2019controllable}, offer end-to-end architectures but are limited to specific domains. Autoregressive models like DALL-E ~\cite{ramesh2021zero} use transformers and large-scale datasets for open-domain generation.
Recently, diffusion models become state-of-the-art in T2I generation. GLIDE \cite{nichol2021glide} innovatively employs a conditional diffusion model to guide the noise diffusion process of images with text. By further introducing image latent features, Stable Diffusion ~\cite{rombach2022high} overcomes resolution constraints and generates amazing high-res results.
Diffusion models also support flexible text-guided image editing and customization. By embedding images and text into a unified latent space, diffusion models can generate customized images of specific concepts based on limited user-provided images and text prompts as guidance \cite{gal2022image,ruiz2022dreambooth,kumari2022multi}. 
However, these diffusion methods are primarily designed for image customization, and a further vectorization step is required to convert them into vector graphics.

\textbf{Image Vectorization.}
Image vectorization aims at representing an image using parametric shape primitives, such as control points and color parameters of vector paths~\cite{quint2003scalable}. Traditional vectorization methods follow a pipeline that involves segmenting images into regions based on color similarity ~\cite{kopf2011depixelizing} and fitting curves to the region boundaries ~\cite{selinger2003potrace, yang2015effective, favreau2017photo2clipart}. 
Furthermore, Perception ~\cite{hoshyari2018perception} and PolyFit ~\cite{dominici2020polyfit} refine the curve-fitting process to better align with perceptions.
However, these traditional vectorization approaches ignore the topological information inherent in raster images and are thus incapable of obtaining layer-wise paths in SVGs.
The advent of differentiable rendering techniques ~\cite{li2020differentiable} has improved image vectorization by enabling the optimization of path parameters using loss functions defined in image space. LIVE \cite{ma2022towards} uncovers layer-wise topology by directly optimizing path parameters under the reconstruction supervision of the input image. SketchRNN ~\cite{ha2017neural} combines differentiable rendering with a recurrent neural network (RNN) to predict vector paths in order. Subsequently, Sketchformer ~\cite{ribeiro2020sketchformer} uses a transformer network to recover sketch strokes from raster images. 
Im2vec ~\cite{reddy2021im2vec} introduces a VAE to predict ordered vector paths from raster images. 
Despite their success, inadequate path initialization and optimization can result in messsy and redundant paths.

\textbf{Text-to-Vector Generation.}
Besides combining T2I generation and image vectorization, another line of work for T2V generation is to directly optimize SVG paths guided by pre-trained vision-language models, such as the CLIP model ~\cite{radford2021learning} or the diffusion model ~\cite{rombach2022high}. 
CLIPDraw ~\cite{frans2022clipdraw} optimizes the image-text similarity metric in CLIP latent space to generate vector graphics from text prompts.
StyleCLIPDraw ~\cite{schaldenbrand2022styleclipdraw} augments CLIPDraw by introducing a style loss, enabling both artistic and textual control over the synthesized drawings.
ES-CLIP ~\cite{tian2022modern} adopts an evolution strategy to optimize triangle-based vector elements. To vectorize and edit the input image based on text prompts, CLIPVG \cite{song2022clipvg} progressively enhances SVG details by adding additional paths. Building upon the strong visual and semantic priors of pre-trained diffusion models, VectorFusion \cite{jain2022vectorfusion} optimizes an SVG consistent with a text prompt using Score Distillation Sampling (SDS) loss.
However, challenges persist in applying these methods to vector graphics customization. Ensuring that the resulting SVG maintains the characteristics of the exemplar SVG, even with initialization, can be difficult due to diverse optimization processes. Without regularization, resulting SVGs may have messy and stacked paths, complicating further editing and modification.

\section{Overview}
\label{sec:overview}

Given an exemplar SVG and a text prompt, our goal is to generate a customized SVG that conforms to the semantics of the given text prompt while preserving the visual identity of the exemplar SVG. Additionally, the customized SVG should maintain the valid path properties and layer relationships of the exemplar SVG. Since an SVG is composed of a set of paths, we define each path as a piecewise Bézier spline composed of several cubic Bézier curves connected end-to-end and filled with a uniform color $c$. Therefore, we can represent a path as $Path=(p_1,p_2,\ldots,p_d,c)$, where $\{p_j\}_{j=1}^d$ are $d$ control points used to define the cubic Bézier curves. Given an exemplar SVG consisting of $n$ parametric paths, denoted as $SVG_E =\{Path_1, Path_2,...,Path_n\}$, and a text prompt $T$, our goal is to obtain a customized SVG composed of $m$ optimized paths, denoted as $SVG_C = \{Path_1^*, Path_2^*, \ldots,Path_m^*\}$.

Our framework, as shown in \reffig{pipeline}, is inspired by experts' design process. When designing a series of SVGs for a specific concept, designers initially select essential semantic elements to reuse across SVGs, place them in appropriate regions, and progressively adjust the paths concerning their position, shape, color, and hierarchical order to generate a set of SVGs that share the same visual identity but have different contexts~\cite{ambrose2019fundamentals}. Inspired by this process, our text-guided vector graphics customization framework can be decomposed into three stages: concept fine-tuning, path alignment, and path optimization.

\noindent \textbf{Concept Fine-tuning} (\refsec{concept_fine_tuning}).
To avoid the need for training on large vector graphics datasets, our first step is to leverage the powerful visual and semantic priors induced by a pre-trained T2I generation model for image customization. Specifically, we use a differentiable rasterizer \cite{li2020differentiable}, denoted as $\mathcal{R}$, to obtain a rasterized exemplar SVG, denoted as $\mathcal{I}_E=\mathcal{R}(SVG_E)$. We then fine-tune the pre-trained Stable Diffusion model \cite{rombach2022high} based on the exemplar image $\mathcal{I}_E$ to generate a customized raster image, denoted as $\mathcal{I}_C$, that reflect the desired modifications indicated by the text prompt $T$ (see \reffig{pipeline} (b)).

\noindent \textbf{Path Alignment} (\refsec{path_alignment}).
To preserve path properties and layer-wise information from the exemplar SVG, we use pre-trained DINO-ViT features \cite{caron2021emerging} to establish semantic correspondences between the exemplar image $\mathcal{I}_E$ and the customized image $\mathcal{I}_C$. We then transform reusable paths from $SVG_E$ to their corresponding connected component in $\mathcal{I}_C$, adjusting their positions and scales as necessary. For unmatched components in $\mathcal{I}_C$, we add new paths. As a result, we obtain $SVG_C^0 = \{Path_1^0, Path_2^0,..., Path_m^0\}$ as an initialization (see \reffig{pipeline} (c)) for the customized SVG.

\noindent \textbf{Path Optimization} (\refsec{path_optimization}).
After initialization, we optimize the geometric and color parameters of paths based on both image-level loss and vector-level loss. The differentiable rasterizer connects the path parameters $SVG_{C}$ and the customized image $\mathcal{I}_C$. Path optimization refines the path parameters to more accurately fit $\mathcal{I}_C$, while maintaining smooth path deformation. This process ultimately yields the customized SVG with optimized paths $SVG_C = \{Path_1^*, Path_2^*,...,Path_m^*\}$ (see \reffig{pipeline} (d)).

\section{Method}
\label{sec:method}

\subsection{Concept Fine-tuning} 
\label{sec:concept_fine_tuning}

Instead of training a T2V model directly for SVG generation and editing, which requires a large dataset of vector graphics that may not be readily available, we utilize a pre-trained T2I generation model like Stable Diffusion \cite{rombach2022high}. We first generate a customized image using this model and then proceed to vectorize it. However, since the Stable Diffusion model is solely trained on text prompts, it cannot maintain the visual identity of our exemplar SVG. To overcome this limitation, we introduce a Concept Fine-tuning approach, where the model is fine-tuned on the rasterized exemplar SVG. This fine-tuning enables the model to generate customized images that reflect the desired modifications specified by the text prompt while preserving the visual identity of the exemplar.

Given an exemplar image $\mathcal{I}_E$, which is rendered from $SVG_E$, we aim to fine-tune the Stable Diffusion model to generate an image that resembles $\mathcal{I}_E$ under a specific text prompt denoted as 'A clipart of $V*$'. $V*$ is initialized with a rarely occurring token embedding and optimized concurrently with the fine-tuning. However, fine-tuning all the weights of the Stable Diffusion model may lead to overfitting on $\mathcal{I}_E$, resulting in reduced editing variations. Thus, we adopt an effective concept fine-tuning strategy that only fine-tunes the cross-attention blocks between text and image, following the approach in \cite{kumari2022multi}. 
Specifically, latent image features form the query, while text features are projected into the key and value using the projection matrices $W_k$ and $W_v$.
The cross-attention operation is conducted between the query, key and value, and we only fine-tune the parameters $W_k$ and $W_v$, as shown in \reffig{pipeline} (b).

We fine-tune the Stable Diffusion model "SD-V1-5" on $\mathcal{I}_E$, then we can concatenate $V*$ with the input text prompt $T$ to form a new prompt such as "$V*$ running, holding a trophy," to generate a customized raster image $\mathcal{I}_C$ that reflects the intended edits.
To further promote the generation of customized images with flat vector style, we add a suffix to the prompt during inference: "minimal flat 2d vector. no outlines. trending on artstation." As sampling results may not always align well with text prompts, we generate $K$ images and select the sample that exhibits the highest consistency with the text prompt based on similarities in CLIP ViT-B/32 features \cite{radford2021learning}. For generated images containing backgrounds, we use U2-Net \cite{qin2020u2} to automatically mask out background.

\subsection{Path Alignment}
\label{sec:path_alignment}

After obtaining a customized image $\mathcal{I}_C$, our goal is to convert it into a vectorized version $SVG_C$. However, the quality of image vectorization strongly relies on initialization. Random initialization often leads to unsuccessful topological extraction and generates redundant shapes. To address this issue, LIVE \cite{ma2022towards} attempts to determine the initial path location based on the color and size of each connected component. However, color and size features are inadequate for retrieving topological path information due to the complex occlusion relationships among elements.

To tackle this challenge, our proposed solution is to reuse suitable paths from the exemplar $SVG_E$ as initialization. To accomplish this, we need to establish correspondences between paths in $SVG_E$ and connected components in $\mathcal{I}_C$. We can then transform and scale selected paths of $SVG_E$ to match their corresponding components in $\mathcal{I}_C$. As customization may require appearance changes, we use semantic features from the pre-trained deep DINO-ViT model "dino-vitb8" \cite{caron2021emerging} instead of relying on lower-level features such as color and shape for matching. This approach enables us to establish more reliable semantic correspondences.

\begin{figure}[tbp]
  \centering
  \includegraphics[width=1.0\columnwidth]{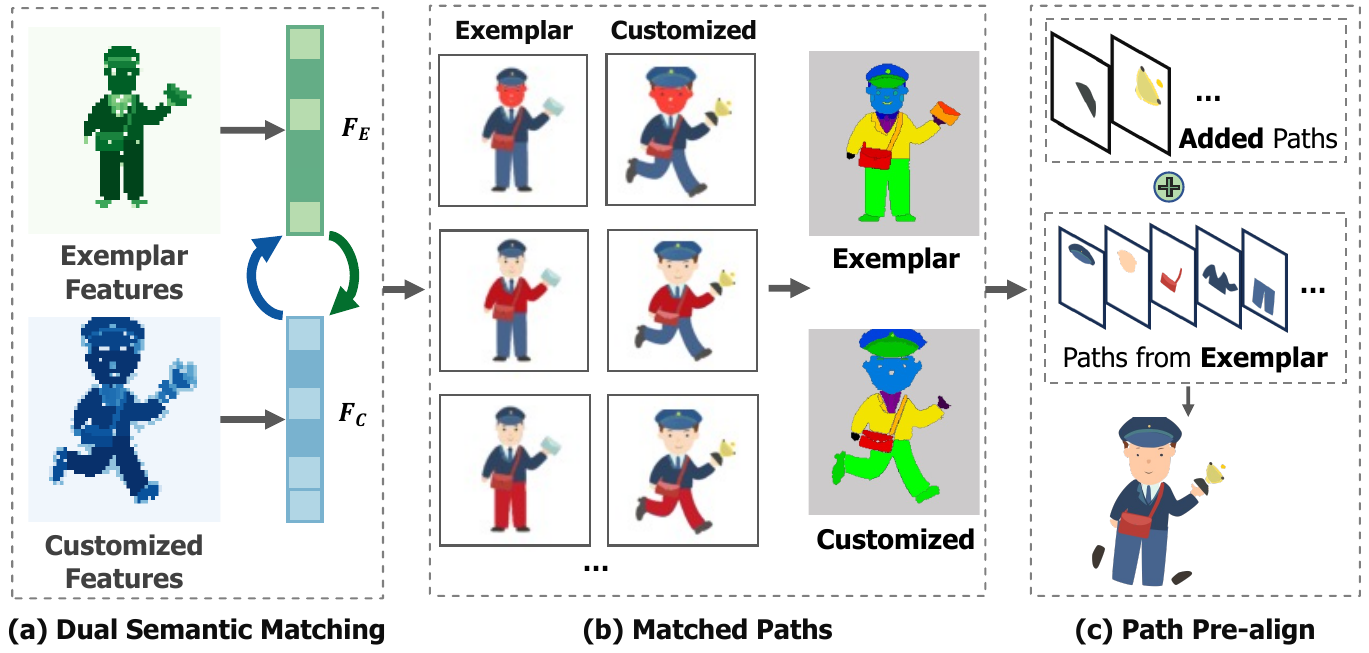}
  \caption{\label{fig:path_alignment_dino} Our path alignment module for generating initial paths for customized SVG based on deep DINO-ViT features.}
\end{figure}

\textbf{Dual Semantic Matching.}
Deep DINO-ViT features have been explored and applied to various visual tasks, such as image co-segmentation and object part correspondence~\cite{amir2021deep, hadjivelichkov2023one}, providing fine-grained local semantic features of object parts.
Therefore, we adopt DINO-ViT features to represent paths in $SVG_E$ and components in $\mathcal{I}_C$ in order to find semantic correspondences between them.
In DINO-ViT, each layer contains a set of features, including a query, key, value, and token. Following the analysis in DeepViT~\cite{amir2021deep}, we utilize the key features to represent the semantic information of images. Therefore, both $\mathcal{I}_E$ and $\mathcal{I}_C$ are fed into a pre-trained DINO-ViT model to obtain the corresponding key feature maps $F_E$ and $F_C$, where each point in the feature map represents semantic feature of an $8\times8$ image patch, as illustrated in \reffig{path_alignment_dino} (a).

To perform semantic matching, we compute corresponding features for each path. Specifically, we segment $\mathcal{I}_E$ according to the paths in $SVG_E$, where each segment corresponds to a path with no overlaps. We then down-sample the segmentation map with the same resolution as $F_E$ and calculate the average features of $F_E$ within each segment. This process generates the feature representation for the $n$ paths, forming a path feature vector $\{F_{Path_i}\}_{i=1}^n$. 
Similarly, to obtain a feature vector for the $m$ components in $\mathcal{I}_C$, we segment the image into connected components with a uniformly-filled color and accumulate feature points in $F_C$. This feature vector is denoted as $\{F_{Comp_j}\}_{j=1}^m$.
We then compute the pairwise cosine similarity between the feature vectors $F_{Path_{i}}$ and $F_{Comp_{j}}$ to construct a similarity matrix $Sim$, where $Sim(i,j)$ denotes the similarity between the $i$-th path and the $j$-th component.
To build robust correspondence between paths and components, we perform dual semantic matching. This involves applying a dual softmax to $Sim$ by normalizing it in both row and column dimensions, resulting in a dual similarity matrix $Sim_{2}$ as follows:
\begin{equation}
  Sim_{2} = \text{softmax}(Sim, dim=row) \odot  \text{softmax} (Sim, dim=col)
\end{equation}
where $\odot$ indicates the element-wise product.

As shown in \reffig{path_alignment_dino} (b), for each component with index $j$, we match it with the $i$-th path if ${{Sim}_{2}}(i,j)$ is the largest value in the $j$-th column. However, we only consider the match to be valid if ${{Sim}_{2}}(i,j)$ exceeds a predefined threshold $\tau_{th}=0.0625$. Scores falling below this threshold indicate low similarity and are discarded.
By leveraging the dual semantic matching mechanism, we can obtain more accurate semantic correspondence.

\textbf{Path Pre-align.}
As shown in \reffig{path_alignment_dino} (c), to initialize paths that align with $\mathcal{I}_C$, we transform and scale the matched paths from $SVG_E$ to their corresponding components in $\mathcal{I}_C$. 
Specifically, we compute the convex hull for matched image components that have a corresponding SVG path, estimate the affine transformation matrix between the hull and path boundary points using the coherent point drift (CPD) algorithm ~\cite{myronenko2010point}, and accordingly transform the path.
For unmatched components in $\mathcal{I}_C$, we use a curve fitting method~\cite{selinger2003potrace} to fit their boundaries with new paths composed of piecewise cubic Bézier curves. This results in a new path set $SVG_C^0 = \{Path^0_1, Path^0_2,\ldots, Path^0_m\}$, which consists of the matched and transformed paths from the exemplar SVG, along with the fitted paths for the unmatched components.

\subsection{Path Optimization}
\label{sec:path_optimization}

\begin{figure}[tbp]
  \centering
  \includegraphics[width=0.85\columnwidth]{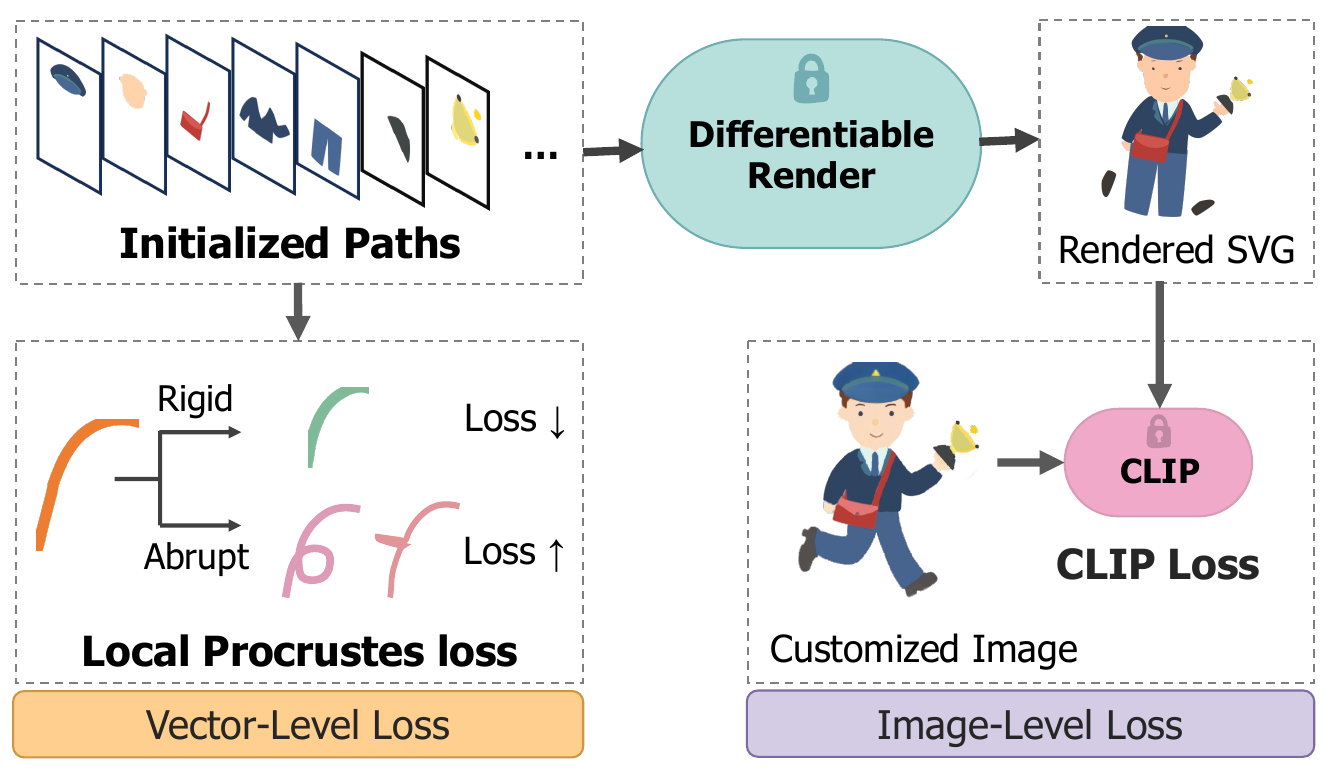}
  \caption{\label{fig:path_optm_loss} The path optimization module of our method. }
\end{figure}

The path alignment step provides a good initialization for the customized vector graphic, which does not need to be highly accurate. To better reconstruct the customized image with our customized SVG, we perform a path optimization process. Given the initial path parameters $SVG_C^0$, the goal is to obtain a refined set of path parameters $SVG_C$ by optimizing control point positions and fill colors to better align with $\mathcal{I}_C$ while preserving the regularity of the initial paths. To accomplish this, we employ both image-level and vector-level losses to guide the path optimization process (\reffig{path_optm_loss}).

\textbf{Image-Level Loss.}
The differentiable rasterizer ~\cite{li2020differentiable} establishes a connection between the SVG parameters and the raster image, enabling backpropagation of image-level losses to optimize path parameters such as control point positions and filling colors. Drawing inspiration from CLIPasso ~\cite{vinker2022clipasso}, we employ a CLIP-based loss to evaluate the image-level similarity between the rendered image of the customized SVG $SVG_C^i$ at the current $i$-th iteration and the target image $\mathcal{I}_C$. We compute the $L_2$ distance between their intermediate-level activations of CLIP as follows:
\begin{equation}
L_{CLIP} = \sum_{l} \lVert CLIP_l(\mathcal{I}_C) - CLIP_l(\mathcal{R}(SVG_C^i)) \rVert _2^2,
\end{equation}
where $CLIP_l$ denotes the CLIP encoder activation at layer $l$. For this purpose, we utilize layers 3 and 4 of the ResNet101 CLIP model. By employing the CLIP-based loss, we encourage the customized SVG to be faithful to the customized image.

\textbf{Vector-Level Loss.}
To avoid messy paths during optimization while still allowing for necessary deformation, we incorporate the Procrustes distance ~\cite{wang2008manifold} into our optimization process. The Procrustes distance $\mathrm{ProDist} (P_1,P_2)$ measures the similarity between two point sets $P_1$ and $P_2$ by minimizing their dissimilarity through a sequence of rigid transformations, including translation, scaling, and rotation operations. After applying these rigid transformations, the Procrustes distance is computed by summing up the Euclidean distances between the corresponding points in the aligned two point sets.

We apply the Procrustes distance to a local window of every control point for every path. Let $p^i_{j,k}$ denote the $k$-th control point in the $j$-th path of the optimizing $SVG_C^i$ in the current $i$-th iteration, while $p^0_{j,k}$ denotes the same control point in the initialized $SVG_C^0$. Our local Procrustes loss is defined as:
\begin{equation}
  L_{Procrustes} = \sum_{j}^m \sum_{k}^{d_j} \mathrm{ProDist}(\mathcal{W}(p^0_{j,k}), \mathcal{W}(p^i_{j,k})).
\end{equation}
Here, $\mathcal{W}(p)$ denotes a neighboring point set with a local window centered at the control point $p$. The window size can be adjusted to balance local and global rigidity, with a larger window size implying a stronger constraint on global rigidity. In our experiments, we set the window size to be 4 neighboring control points. Procrustes distances are accumulated for all $d_j$ control points, where $j$ ranges from 1 to $m$, the total number of paths.

This local Procrustes loss applies geometric constraints to control points, regulating the shape deformation of paths to be as locally rigid as possible. As illustrated in Fig. 4, the local Procrustes loss is small when the shape deformation remains locally rigid, while it becomes larger in the presence of curve intersections or abrupt curvature changes. By constraining path optimization in this manner, we preserve the regularity of the paths while allowing for some degree of deformation to match the desired customization.

The overall loss function $L$ is defined as a combination of both CLIP loss and local Procrustes loss:
\begin{equation}
\mathcal{L}=\mathcal{L}_\text{CLIP}+\lambda\mathcal{L}_\text{Procrustes}
\end{equation}
where $\lambda$ is a balance factor. As there is a trade-off between the CLIP loss and the local Procrustes loss, we prioritize the CLIP loss in the early stages of optimization to achieve a better fit to the customized image, and then increase the importance of the local Procrustes loss to regularize the path shape. Specifically, we linearly increase $\lambda$ from $0.01$ to $0.04$ over 200 iterations. After convergence, we obtain the optimized customized SVG, denoted as $SVG_C = \{Path^*_1, Path^*_2,...,Path^*_m\}$.

\section{Experiment}
\label{sec:experiment}

\textbf{Experiment Setup.} 
To evaluate our method, we collected 100 vector graphics from \textit{freepik}\footnote{https://www.freepik.com/} and \textit{iconfont}\footnote{https://icofont.com/}, including 45 characters, 35 animals, and 20 scenes.
We select text prompts from the Stable Diffusion Prompts dataset~\cite{dehouche2023text}, which includes diverse actions and scenes. For each SVG, we randomly select 5 prompts and generate corresponding customized SVGs, resulting in 500 pairs of exemplar and customized vector graphics. The average number of paths in our customized SVGs is 28.

\subsection{Evaluation Metrics}
\label{sec:evaluation_metrics}

To assess the quality of text-to-SVG synthesis, we evaluate the generated SVGs at vector-level, image-level and text-level.

\textbf{Vector-level Evaluation}
Evaluating text-to-SVG synthesis is challenging without ground truth SVGs. We directly compare the customized SVG paths with the input exemplar SVG paths based on shape similarity, since the exemplar SVGs can be regarded as well-designed SVGs.
Higher similarity indicates that the customized SVG better preserves the original shape properties.
In addition, we evaluate vectorization quality based on smoothness criteria derived from the prior perception and computer graphics research ~\cite{dominici2020polyfit}.
Specifically, the vector-level evaluation metrics include:
(a) \textbf{Shape similarity ($Sim_{shape}$)}: $1-D_{H}$, where $D_{H}$ is Hausdorff distance between path control points of the exemplar SVG and the customized SVG.
(b) \textbf{Smoothness}: Inverse of the average curvature variation of the paths in customized vector graphics.

\textbf{Image-level Evaluation}
As the customized image should keep the same visual identity as the exemplar image, and the customized SVG should follow the customized image after rendering, we define the image-level metrics as:
(a) \textbf{Exemplar image similarity ($Sim_{exp}$)}: Cosine similarity between the exemplar image and the rendered customized SVG result in CLIP space ~\cite{vinker2022clipasso}.
(b) \textbf{Customized image similarity ($Sim_{cus}$)}: $1-MSE$, where $MSE$ is the mean squared error between the customized image and the rendered customized SVG result in RGB space.

\textbf{Text-level Evaluation}
To compute whether the customized SVG is aligned with the input text prompt, we define the text-level similarity by calculating the CLIP cosine simlairty between the text prompt and rendered customized SVG (\textbf{$Sim_{CLIP}$}).

\subsection{Comparison with Existing Methods}
\label{sec:comparison_with_existing}

\begin{figure}[tbp]
  \centering
  \includegraphics[width=1.0\columnwidth]{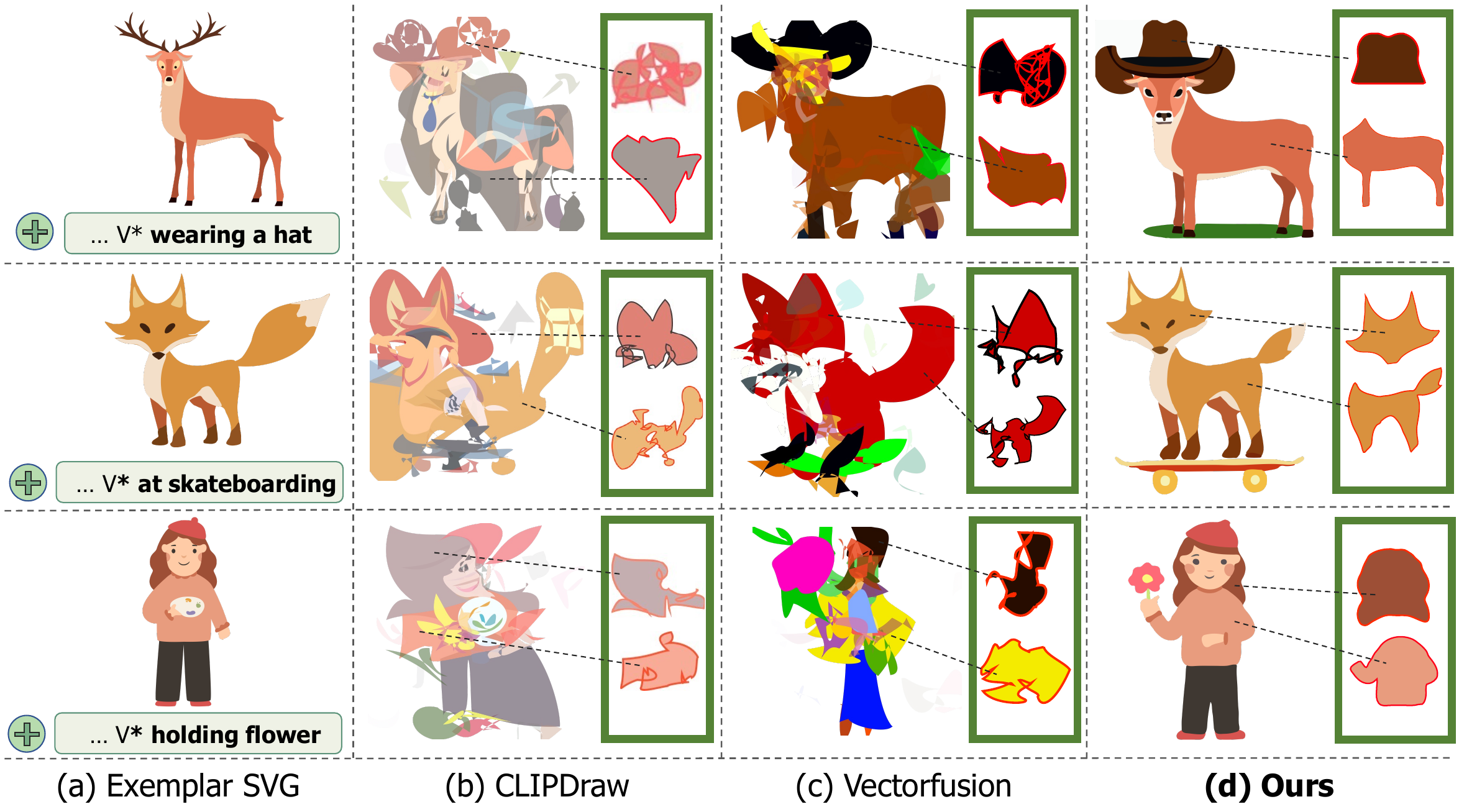}
  \caption{\label{fig:result_vecfusion} Qualitative comparison with text-guided SVG optimization methods. Exemplar SVGs: the $1^{st}$ row is from \copyright{Freepik}; the $2^{nd}$ row is from illustac creator \copyright{Zarame}; the $3^{rd}$ row is from envatoelements creator \copyright{Telllu}.}
\end{figure}

We compare our proposed pipeline with two types of T2V works: vectorization with T2I and text-guided SVG optimization methods. For \textbf{vectorization with T2I} methods, we use the same customized images generated by our concept fine-tuning method, and select both traditional and deep learning-based vectorization methods for comparisons:
(a) \textbf{Potrace} ~\cite{selinger2003potrace}: A traditional vectorization method that segments images into regions and then fits piecewise cubic Bézier curves to the region boundaries.
(b) \textbf{LIVE} ~\cite{ma2022towards}: A deep learning method generates SVGs by initializing path locations based on region colors and then uses loss functions to optimize paths. We use the same number of paths as our method for fairness, and set the control points of each path to 12, consistent with their paper.
For \textbf{text-guided SVG optimization}, we compare our method with CLIP-based and diffusion-based optimization approaches:
(c) \textbf{CLIPDraw} ~\cite{frans2022clipdraw}: This method optimizes CLIP's image-text similarity metric to generate vector graphics from text prompts.
(d) \textbf{VectorFusion} ~\cite{jain2022vectorfusion}: This approach employs SDS loss to optimize SVGs to be consistent with given text prompts.
We use the exemplar SVG as initialization. For a fair comparison, if the exemplar SVG has fewer paths than our initialization, we randomly add paths to match the number of paths as our method.
The quantitative result is shown in \reftab{table_quality_eval} and the qualitative results are shown in \reffig{result_vecfusion} and \reffig{result}.

\begin{table}[tbp]
  \caption{Quantitative comparison with existing methods. $Sim_{s}$, $Sim_{cus}$ and $Sim_{exp}$ denote shape similarity between $SVG_E$ and $SVG_C$, RGB similarity between $\mathcal{I}_C$ and rendered $SVG_C$, CLIP similarity between $\mathcal{I}_E$ and rendered $SVG_C$.}
  \small
  \resizebox{\linewidth}{!}{
  \begin{tabular}{|c|cc|cc|c|}
    \hline
    \multirow{2}{*}{Methods} & \multicolumn{2}{c|}{Vector Level}                                                                                       & \multicolumn{2}{c|}{Image Level}                                                       & Text level                 \\ \cline{2-6} 
                             & \multicolumn{1}{c|}{$Sim_{s}$ $\uparrow$} & \multicolumn{1}{c|}{Smooth $\uparrow$} & \multicolumn{1}{c|}{$Sim_{cus}$ $\uparrow$} & $Sim_{exp}$ $\uparrow$ & $Sim_{CLIP}$ $\uparrow$ \\ \hline
    Potrace                  & \multicolumn{1}{c|}{-}                           & \multicolumn{1}{c|}{0.7919}              & \multicolumn{1}{c|}{0.9937}                           & 0.9076                         & 0.3218                     \\ \hline
    LIVE                     & \multicolumn{1}{c|}{-}                           & \multicolumn{1}{c|}{0.5929}              & \multicolumn{1}{c|}{0.9906}                           & 0.8576                         & 0.2859                     \\ \hline
    CLIPDraw                 & \multicolumn{1}{c|}{0.1380}                      & \multicolumn{1}{c|}{0.4613}              & \multicolumn{1}{c|}{-}                                & 0.7862                         & 0.3175                     \\ \hline
    Vectorfusion             & \multicolumn{1}{c|}{0.1434}                      & \multicolumn{1}{c|}{0.4895}              & \multicolumn{1}{c|}{-}                                & 0.7214                         & 0.2718                     \\ \hline
    \textbf{Ours}                     & \multicolumn{1}{c|}{\textbf{0.8121}}                      & \multicolumn{1}{c|}{\textbf{0.7931}}              & \multicolumn{1}{c|}{\textbf{0.9944}}                           & \textbf{0.9111}                         & \textbf{0.3227}                     \\ \hline
  \end{tabular}}
  \label{tab:table_quality_eval}

\end{table}

\textbf{Vectorization with T2I Methods.}
\textbf{Potrace} ~\cite{selinger2003potrace} segments images into regions based solely on color, thus it can generate multiple redundant paths for a single semantic element, such as multiple paths for a single pants in the $3^{rd}$ row of \reffig{result} (c). This leads to higher complexity. It also loses the layer-wise relationships provided by the exemplar SVG and curve smoothness along boundaries due to the lack of pixel-level supervision.
As for \textbf{LIVE} ~\cite{ma2022towards}, due to the path complexity caused by the occlusion, only relying on color and size features to reconstruct the customized image fails to maintain the topological path properties, such as the broken paths in \reffig{result} (d). When using the same number of paths as our method, the similarity to the customized image reduces a lot as shown in \reftab{table_quality_eval}. 
In contrast, our model employs a semantic-based path alignment module that preserves layer-wise path properties and a path optimization module to obtain smooth and accurate final paths, outperforming these two methods quantitatively.

\textbf{Text-guided SVG Optimization Methods.}
Although these methods directly inherit the original layer-wise properties, neither the CLIP loss nor the SDS loss is sufficient for preserving path quality. The original paths may undergo complex transformations to generate text-conforming SVGs, resulting in low shape similarity with the exemplar SVG (\reftab{table_quality_eval}). As shown in \reffig{result_vecfusion}, there are instances of intersecting paths and abrupt curvature changes, which lead to diminished path smoothness (\reftab{table_quality_eval}) and visually unappealing outcomes.
While CLIPDraw directly optimizes CLIP loss and achieves high CLIP metrics, the rendered SVGs appear messy.
Our pipeline generates much clearer vector graphics than CLIPDraw and VectorFusion, by finetuning the diffusion model as a prior for customized images and preserving the desirable path properties of the exemplar SVG during the vectorization process.

\subsection{User Study}
\label{sec:user_study}

\begin{figure}[tbp]
  \centering
  \includegraphics[width=\columnwidth]{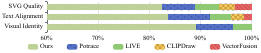}
  \caption{\label{fig:user_study_bar} User Study. We show the human preferences in \%.}
\end{figure}

We conduct a perceptual study to evaluate our vector graphics customization from three perspectives: overall SVG quality, alignment with text prompt, and similarity in visual identity with the exemplar SVG. We randomly select 12 pairs of exemplar SVGs and text prompts from our test set, and generate customized SVGs using the baseline methods in \refsec{comparison_with_existing} and our approach.
In each question, we display the results of different methods in random order and ask 30 participants to select the best result among five options for each evaluation metric. 
\reffig{user_study_bar} demonstrates the superior performance of our method, as it achieves the highest preference in all evaluation metrics. Specifically, our method obtains 82.5\% of votes for overall SVG quality, 83.7\% for text alignment, and 89.1\% for similarity in visual identity. The results show the effectiveness of our method in generating high-quality customized SVGs based on text prompts while preserving the visual identity of the exemplar SVG.

\subsection{Ablation Study}
\label{sec:ablation_study}

\begin{figure}[tbp]
  \centering
  \includegraphics[width=\columnwidth]{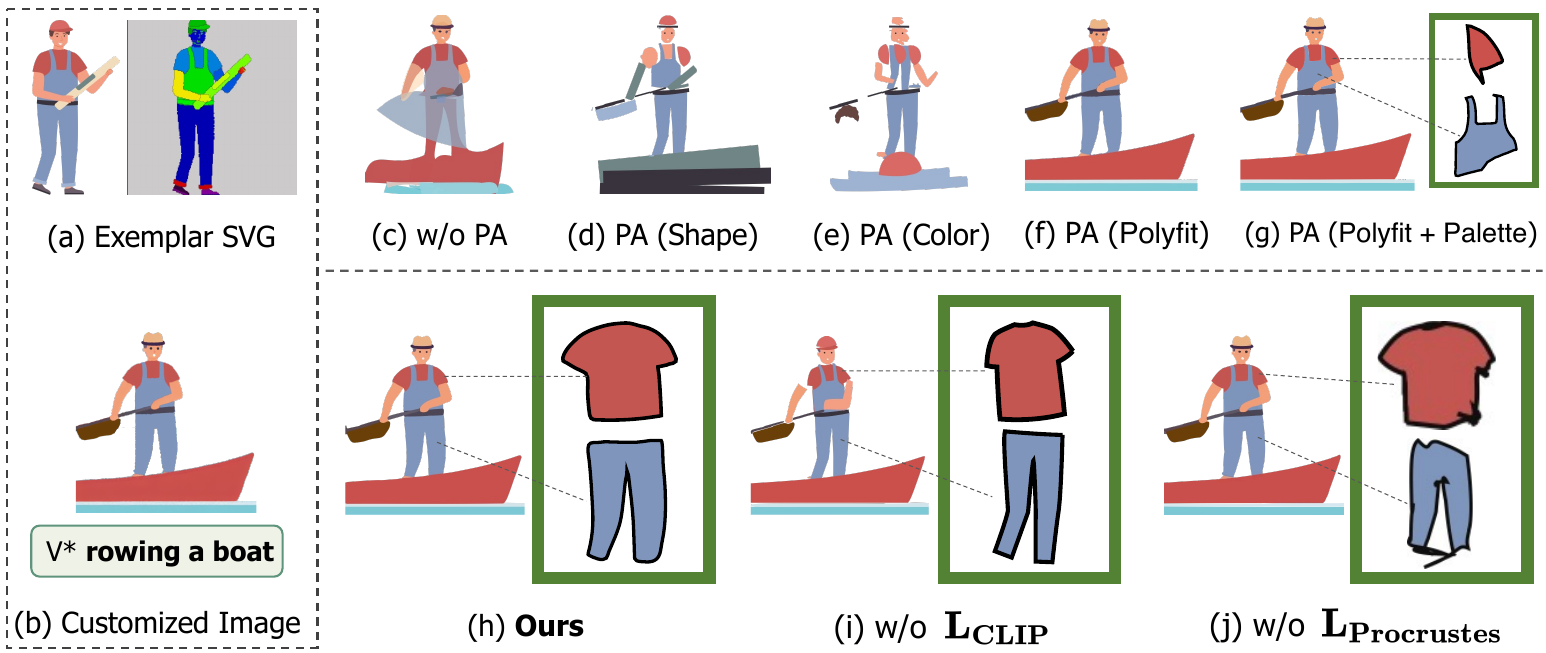}
  \caption{\label{fig:ablation} Qualitative results on ablation study. The exemplar SVG is from Freepik creator \copyright{Alexdndz}.}
\end{figure}

\textbf{The Effectiveness of Path Alignment.}
To investigate the effectiveness of our path alignment module, we compare it with five baselines. The first one is using the input exemplar SVG directly as the path initialization (\textbf{w/o PA}). Due to the lack of the path alignment module, with the large visual difference between the customized image and the rendered exemplar SVG, significant distortions occur after path optimization, as shown in \reffig{ablation} (c).
Rather than using our semantic features for finding the matched paths, the other two baselines are using \textbf{Color} and \textbf{Shape} similarities, respectively. For color, we directly compute the MSE in RGB space and for shape, we compute the distance after normalizing paths under rigid transformations. Given that many paths share similar colors and the large structure/shape deformation exists in the customized image, using shape and color similarities leads to incorrect path matches as shown in \reffig{ablation} (d, e). Thus the final optimized paths become messy (\reftab{ablation_study}).
We also ablate with standard vectorization methods: PolyFit~\cite{dominici2020polyfit} and PolyFit with Palette compaction ~\cite{yang2023subpixel} to get initial paths. The latter method reduces noisy SVG paths by compacting the color palette of the customized image.
Although their initial vectorizations fit images well, the topological problem remains, as image segmentation cannot reflect occlusion relationships (\reffig{ablation} (f, g)). Instead, our model derives layer-wise paths from the exemplar SVG, preserving favorable path properties.


\noindent\textbf{The Effectiveness of Path Optimization.}
We ablate the loss design by individually removing each of the CLIP loss and local Procrustes loss in the path optimization module. As shown in \reffig{ablation} (i), when the CLIP loss is removed (\textbf{w/o $L_{CLIP}$}), the optimization retains initialization and fails to deform paths to align with the customized image, resulting in low customized image similarity ($Sim_{cus}$ in \reftab{ablation_study}). When the local Procrustes loss is removed (\textbf{w/o $L_{Procrustes}$}), the deformed paths suffer from intersection issues and abrupt curvature changes, generating SVG paths of poor quality with low smoothness. Furthermore, the reduced shape similarity ($Sim_{s}$ in \reftab{ablation_study}) suggests that image-level loss alone is insufficient to preserve the shape properties inherited from the exemplar SVG.

\begin{table}[tbp]
  \small
  \caption{Ablation study on path alignment and path optimization modules.}
  \begin{tabular}{|cl|c|c|c|}
  \hline
  \multicolumn{2}{|c|}{Methods}                                                                                                                                        & \begin{tabular}[c]{@{}c@{}} $Sim_s$ $\uparrow$ \end{tabular} & Smooth $\uparrow$ & \begin{tabular}[c]{@{}c@{}} $Sim_{cus}$ $\uparrow$ \end{tabular} \\ \hline
  \multicolumn{1}{|c|}{\multirow{5}{*}{\begin{tabular}[c]{@{}c@{}}Path \\ Alignment \\ (PA)\end{tabular}}}    & w/o PA                                                       & 0.2280                                               & 0.6217     & 0.9879                                                    \\ \cline{2-5} 
  \multicolumn{1}{|c|}{}                                                                              & Shape                                                          & 0.7435                                               & 0.7902     & 0.9934                                                    \\ \cline{2-5} 
  \multicolumn{1}{|c|}{}                                                                              & Color                                                          & 0.7271                                               & 0.7304     & 0.9912                                                    \\ \cline{2-5}
  \multicolumn{1}{|c|}{}                                                                          & Polyfit           & -     & 0.7922     & 0.9948     \\ \cline{2-5} 
  \multicolumn{1}{|l|}{}                                                                          &  Polyfit + Palette & -     & 0.7923     & 0.9946     \\ \hline
  \multicolumn{1}{|c|}{\multirow{2}{*}{\begin{tabular}[c]{@{}c@{}}Path \\ Optimization\end{tabular}}} & w/o $L_{CLIP}$                                                  & 0.9998                                               & 0.8078     & 0.7197                                                    \\ \cline{2-5} 
  \multicolumn{1}{|c|}{}                                                                              & \begin{tabular}[c]{@{}c@{}}w/o $L_{Procrustes}$ \end{tabular} & 0.4153                                               & 0.5030     & 0.9951                                                    \\ \hline
  \multicolumn{2}{|c|}{Ours}                                                                                                                                           & 0.8121                                               & 0.7931     & 0.9944                                                    \\ \hline
  \end{tabular}

  \label{tab:ablation_study}

\end{table}

\section{Conclusion}
\label{sec:conclusion}

\begin{figure}[tbp]
  \centering
  \includegraphics[width=1.0\columnwidth]{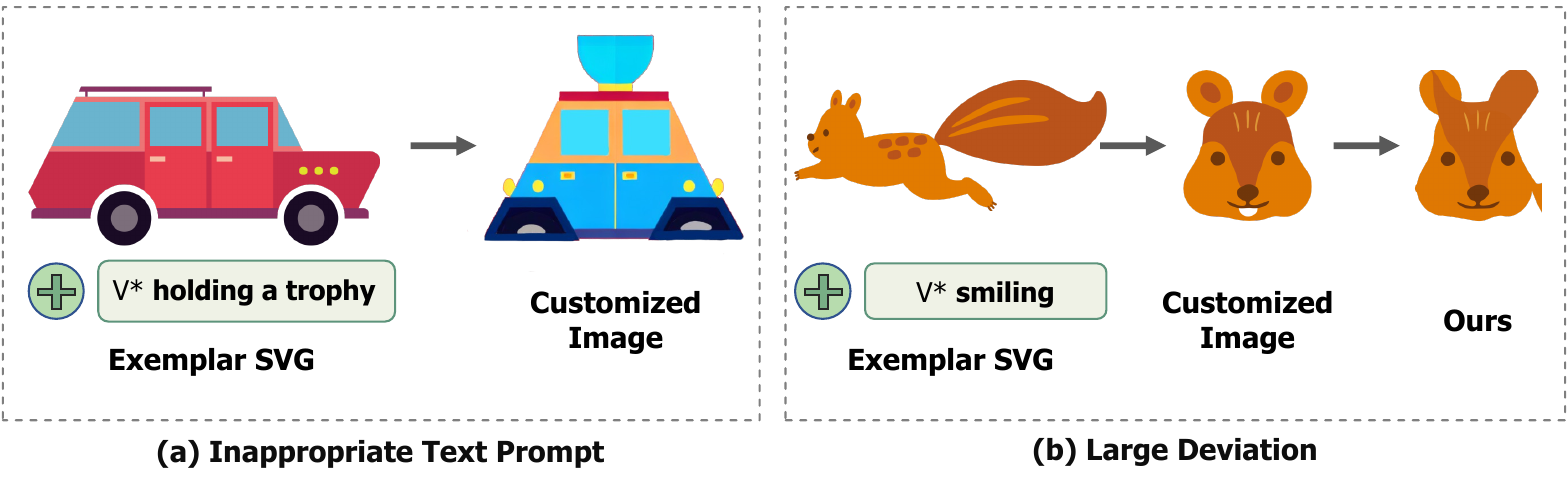}
  \caption{\label{fig:failure_case} Failure cases. Exemplar SVGs: the left is from \copyright{iconfont}; the right is from illustac creator \copyright{Rico}.}
\end{figure}

\begin{figure*}[tbp]
  \centering
  \includegraphics[width=1.0\linewidth]{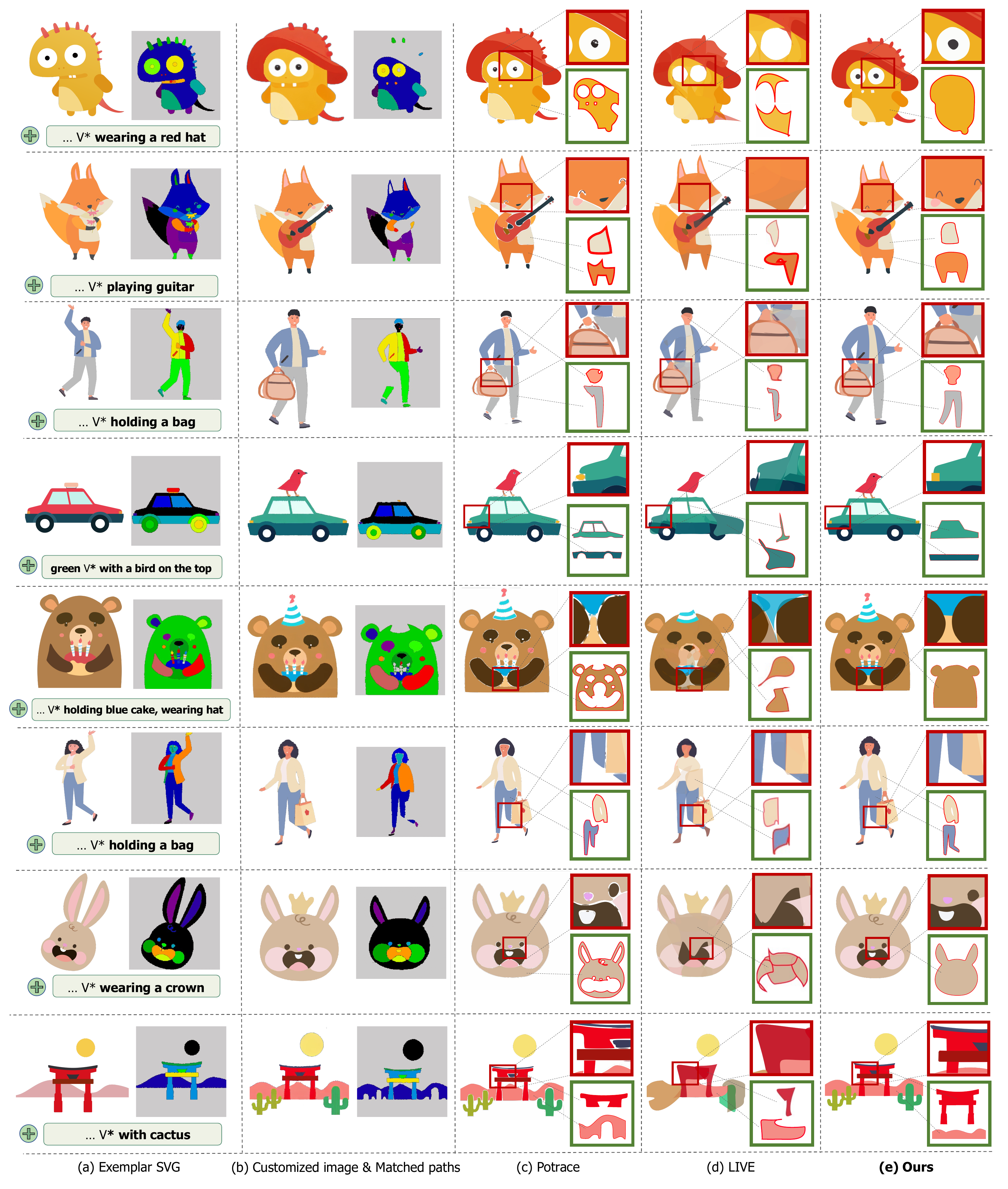}
  \caption{\label{fig:result} Qualitative comparison to vectorization with T2I methods. Exemplar SVGs: the $1^{st}$ and $4^{th}$ rows are from \copyright{iconfont}; the $2^{nd}$ row is from envatoelements creator \copyright{Masastarus}; the $3^{rd}$ and $6^{th}$ rows are from Freepik creator \copyright{Alexdndz}; the $5^{th}$ and $7^{th}$ rows are from \copyright{Freepik}; the $8^{th}$ row is from \copyright{Vectorportal}.}
\end{figure*}

In this paper, we investigate a novel problem called text-guided vector graphics customization, and propose a pipeline that generates customized SVGs based on input text prompts while preserving the visual identity of the input exemplar SVG. Our method reuses paths from the exemplar SVG through newly designed semantic-based path alignment and path optimization modules, maintaining the valid path properties and layer relationships in the final SVGs.
Though our method can achieve high-quality SVG results, we still suffer from failure cases as shown in \reffig{failure_case}.
First, our method relies on the generative capabilities of diffusion model, thus inappropriate pairs of exemplar SVGs and text prompts may result in unreasonable customizations. For example, requesting a car to "hold a trophy" could yield a nonsensical outcome.
Second, if the customized image deviates significantly from the exemplar, the path alignment method may fail to find correct matches. For instance, if the exemplar depicts a side view of a running squirrel, but the customized image shows a frontal view of a smiling squirrel, the path alignment fails to find correct path matches from the exemplar SVG.

\textnormal{\textbf{Ethics.}}
Our work aims to enhance, not replace, the creative flows of vector artists. We acknowledge potential ethical concerns with models trained on internet data (e.g. property rights). Solutions such as content detection systems can help mitigate these risks.

\textnormal{\textbf{Future Work.}}
Our text-guided vector graphics customization exhibits visual creativity. In future work, we plan to integrate human-in-the-loop methods to assist designers in achieving more precise vector graphics generation and editing.

\begin{acks}
The work described in this paper was substantially supported by a GRF grant from the Research Grants Council (RGC) of the Hong Kong Special Administrative Region, China [Project No. CityU 11216122].
\end{acks}

\bibliographystyle{ACM-Reference-Format}
\bibliography{sample-sigconf}


\begin{thebibliography}{37}


\ifx \showCODEN    \undefined \def \showCODEN     #1{\unskip}     \fi
\ifx \showDOI      \undefined \def \showDOI       #1{#1}\fi
\ifx \showISBNx    \undefined \def \showISBNx     #1{\unskip}     \fi
\ifx \showISBNxiii \undefined \def \showISBNxiii  #1{\unskip}     \fi
\ifx \showISSN     \undefined \def \showISSN      #1{\unskip}     \fi
\ifx \showLCCN     \undefined \def \showLCCN      #1{\unskip}     \fi
\ifx \shownote     \undefined \def \shownote      #1{#1}          \fi
\ifx \showarticletitle \undefined \def \showarticletitle #1{#1}   \fi
\ifx \showURL      \undefined \def \showURL       {\relax}        \fi
\providecommand\bibfield[2]{#2}
\providecommand\bibinfo[2]{#2}
\providecommand\natexlab[1]{#1}
\providecommand\showeprint[2][]{arXiv:#2}

\bibitem[Ambrose et~al\mbox{.}(2019)]%
        {ambrose2019fundamentals}
\bibfield{author}{\bibinfo{person}{Gavin Ambrose}, \bibinfo{person}{Paul Harris}, {and} \bibinfo{person}{Nigel Ball}.} \bibinfo{year}{2019}\natexlab{}.
\newblock \bibinfo{booktitle}{\emph{The fundamentals of graphic design}}.
\newblock \bibinfo{publisher}{Bloomsbury Publishing}.
\newblock


\bibitem[Amir et~al\mbox{.}(2021)]%
        {amir2021deep}
\bibfield{author}{\bibinfo{person}{Shir Amir}, \bibinfo{person}{Yossi Gandelsman}, \bibinfo{person}{Shai Bagon}, {and} \bibinfo{person}{Tali Dekel}.} \bibinfo{year}{2021}\natexlab{}.
\newblock \showarticletitle{Deep vit features as dense visual descriptors}.
\newblock \bibinfo{journal}{\emph{arXiv preprint arXiv:2112.05814}} \bibinfo{volume}{2}, \bibinfo{number}{3} (\bibinfo{year}{2021}), \bibinfo{pages}{4}.
\newblock


\bibitem[Caron et~al\mbox{.}(2021)]%
        {caron2021emerging}
\bibfield{author}{\bibinfo{person}{Mathilde Caron}, \bibinfo{person}{Hugo Touvron}, \bibinfo{person}{Ishan Misra}, \bibinfo{person}{Herv{\'e} J{\'e}gou}, \bibinfo{person}{Julien Mairal}, \bibinfo{person}{Piotr Bojanowski}, {and} \bibinfo{person}{Armand Joulin}.} \bibinfo{year}{2021}\natexlab{}.
\newblock \showarticletitle{Emerging properties in self-supervised vision transformers}. In \bibinfo{booktitle}{\emph{Proceedings of the IEEE/CVF international conference on computer vision}}. \bibinfo{pages}{9650--9660}.
\newblock


\bibitem[Dehouche and Kullathida(2023)]%
        {dehouche2023text}
\bibfield{author}{\bibinfo{person}{Nassim Dehouche} {and} \bibinfo{person}{Kullathida}.} \bibinfo{year}{2023}\natexlab{}.
\newblock \showarticletitle{What is in a Text-to-Image Prompt: The Potential of Stable Diffusion in Visual Arts Education}.
\newblock \bibinfo{journal}{\emph{arXiv preprint arXiv:2301.01902}} (\bibinfo{year}{2023}).
\newblock


\bibitem[Diebel(2008)]%
        {diebel2008bayesian}
\bibfield{author}{\bibinfo{person}{James~Richard Diebel}.} \bibinfo{year}{2008}\natexlab{}.
\newblock \bibinfo{booktitle}{\emph{Bayesian Image Vectorization: the probabilistic inversion of vector image rasterization}}.
\newblock \bibinfo{publisher}{Stanford University}.
\newblock


\bibitem[Dominici et~al\mbox{.}(2020)]%
        {dominici2020polyfit}
\bibfield{author}{\bibinfo{person}{Edoardo~Alberto Dominici}, \bibinfo{person}{Nico Schertler}, \bibinfo{person}{Jonathan Griffin}, \bibinfo{person}{Shayan Hoshyari}, \bibinfo{person}{Leonid Sigal}, {and} \bibinfo{person}{Alla Sheffer}.} \bibinfo{year}{2020}\natexlab{}.
\newblock \showarticletitle{Polyfit: Perception-aligned vectorization of raster clip-art via intermediate polygonal fitting}.
\newblock \bibinfo{journal}{\emph{ACM Transactions on Graphics (TOG)}} \bibinfo{volume}{39}, \bibinfo{number}{4} (\bibinfo{year}{2020}), \bibinfo{pages}{77--1}.
\newblock


\bibitem[Favreau et~al\mbox{.}(2017)]%
        {favreau2017photo2clipart}
\bibfield{author}{\bibinfo{person}{Jean-Dominique Favreau}, \bibinfo{person}{Florent Lafarge}, {and} \bibinfo{person}{Adrien Bousseau}.} \bibinfo{year}{2017}\natexlab{}.
\newblock \showarticletitle{Photo2clipart: Image abstraction and vectorization using layered linear gradients}.
\newblock \bibinfo{journal}{\emph{ACM Transactions on Graphics (TOG)}} \bibinfo{volume}{36}, \bibinfo{number}{6} (\bibinfo{year}{2017}), \bibinfo{pages}{1--11}.
\newblock


\bibitem[Frans et~al\mbox{.}(2022)]%
        {frans2022clipdraw}
\bibfield{author}{\bibinfo{person}{Kevin Frans}, \bibinfo{person}{Lisa Soros}, {and} \bibinfo{person}{Olaf Witkowski}.} \bibinfo{year}{2022}\natexlab{}.
\newblock \showarticletitle{Clipdraw: Exploring text-to-drawing synthesis through language-image encoders}.
\newblock \bibinfo{journal}{\emph{Advances in Neural Information Processing Systems}}  \bibinfo{volume}{35} (\bibinfo{year}{2022}), \bibinfo{pages}{5207--5218}.
\newblock


\bibitem[Gal et~al\mbox{.}(2022)]%
        {gal2022image}
\bibfield{author}{\bibinfo{person}{Rinon Gal}, \bibinfo{person}{Yuval Alaluf}, \bibinfo{person}{Yuval Atzmon}, \bibinfo{person}{Or Patashnik}, \bibinfo{person}{Amit~H Bermano}, \bibinfo{person}{Gal Chechik}, {and} \bibinfo{person}{Daniel Cohen-Or}.} \bibinfo{year}{2022}\natexlab{}.
\newblock \showarticletitle{An image is worth one word: Personalizing text-to-image generation using textual inversion}.
\newblock \bibinfo{journal}{\emph{arXiv preprint arXiv:2208.01618}} (\bibinfo{year}{2022}).
\newblock


\bibitem[Ha and Eck(2017)]%
        {ha2017neural}
\bibfield{author}{\bibinfo{person}{David Ha} {and} \bibinfo{person}{Douglas Eck}.} \bibinfo{year}{2017}\natexlab{}.
\newblock \showarticletitle{A neural representation of sketch drawings}.
\newblock \bibinfo{journal}{\emph{arXiv preprint arXiv:1704.03477}} (\bibinfo{year}{2017}).
\newblock


\bibitem[Hadjivelichkov et~al\mbox{.}(2023)]%
        {hadjivelichkov2023one}
\bibfield{author}{\bibinfo{person}{Denis Hadjivelichkov}, \bibinfo{person}{Sicelukwanda Zwane}, \bibinfo{person}{Lourdes Agapito}, \bibinfo{person}{Marc~Peter Deisenroth}, {and} \bibinfo{person}{Dimitrios Kanoulas}.} \bibinfo{year}{2023}\natexlab{}.
\newblock \showarticletitle{One-Shot Transfer of Affordance Regions? AffCorrs!}. In \bibinfo{booktitle}{\emph{Conference on Robot Learning}}. PMLR, \bibinfo{pages}{550--560}.
\newblock


\bibitem[Hoshyari et~al\mbox{.}(2018)]%
        {hoshyari2018perception}
\bibfield{author}{\bibinfo{person}{Shayan Hoshyari}, \bibinfo{person}{Edoardo~Alberto Dominici}, \bibinfo{person}{Alla Sheffer}, \bibinfo{person}{Nathan Carr}, \bibinfo{person}{Zhaowen Wang}, \bibinfo{person}{Duygu Ceylan}, {and} \bibinfo{person}{I-Chao Shen}.} \bibinfo{year}{2018}\natexlab{}.
\newblock \showarticletitle{Perception-driven semi-structured boundary vectorization}.
\newblock \bibinfo{journal}{\emph{ACM Transactions on Graphics (TOG)}} \bibinfo{volume}{37}, \bibinfo{number}{4} (\bibinfo{year}{2018}), \bibinfo{pages}{1--14}.
\newblock


\bibitem[Jain et~al\mbox{.}(2022)]%
        {jain2022vectorfusion}
\bibfield{author}{\bibinfo{person}{Ajay Jain}, \bibinfo{person}{Amber Xie}, {and} \bibinfo{person}{Pieter Abbeel}.} \bibinfo{year}{2022}\natexlab{}.
\newblock \showarticletitle{VectorFusion: Text-to-SVG by Abstracting Pixel-Based Diffusion Models}.
\newblock \bibinfo{journal}{\emph{arXiv preprint arXiv:2211.11319}} (\bibinfo{year}{2022}).
\newblock


\bibitem[Kopf and Lischinski(2011)]%
        {kopf2011depixelizing}
\bibfield{author}{\bibinfo{person}{Johannes Kopf} {and} \bibinfo{person}{Dani Lischinski}.} \bibinfo{year}{2011}\natexlab{}.
\newblock \showarticletitle{Depixelizing pixel art}.
\newblock In \bibinfo{booktitle}{\emph{ACM SIGGRAPH 2011 papers}}. \bibinfo{pages}{1--8}.
\newblock


\bibitem[Kumari et~al\mbox{.}(2022)]%
        {kumari2022multi}
\bibfield{author}{\bibinfo{person}{Nupur Kumari}, \bibinfo{person}{Bingliang Zhang}, \bibinfo{person}{Richard Zhang}, \bibinfo{person}{Eli Shechtman}, {and} \bibinfo{person}{Jun-Yan Zhu}.} \bibinfo{year}{2022}\natexlab{}.
\newblock \showarticletitle{Multi-Concept Customization of Text-to-Image Diffusion}.
\newblock \bibinfo{journal}{\emph{arXiv preprint arXiv:2212.04488}} (\bibinfo{year}{2022}).
\newblock


\bibitem[Li et~al\mbox{.}(2019)]%
        {li2019controllable}
\bibfield{author}{\bibinfo{person}{Bowen Li}, \bibinfo{person}{Xiaojuan Qi}, \bibinfo{person}{Thomas Lukasiewicz}, {and} \bibinfo{person}{Philip Torr}.} \bibinfo{year}{2019}\natexlab{}.
\newblock \showarticletitle{Controllable text-to-image generation}.
\newblock \bibinfo{journal}{\emph{Advances in Neural Information Processing Systems}}  \bibinfo{volume}{32} (\bibinfo{year}{2019}).
\newblock


\bibitem[Li et~al\mbox{.}(2020)]%
        {li2020differentiable}
\bibfield{author}{\bibinfo{person}{Tzu-Mao Li}, \bibinfo{person}{Michal Luk{\'a}{\v{c}}}, \bibinfo{person}{Micha{\"e}l Gharbi}, {and} \bibinfo{person}{Jonathan Ragan-Kelley}.} \bibinfo{year}{2020}\natexlab{}.
\newblock \showarticletitle{Differentiable vector graphics rasterization for editing and learning}.
\newblock \bibinfo{journal}{\emph{ACM Transactions on Graphics (TOG)}} \bibinfo{volume}{39}, \bibinfo{number}{6} (\bibinfo{year}{2020}), \bibinfo{pages}{1--15}.
\newblock


\bibitem[Ma et~al\mbox{.}(2022)]%
        {ma2022towards}
\bibfield{author}{\bibinfo{person}{Xu Ma}, \bibinfo{person}{Yuqian Zhou}, \bibinfo{person}{Xingqian Xu}, \bibinfo{person}{Bin Sun}, \bibinfo{person}{Valerii Filev}, \bibinfo{person}{Nikita Orlov}, \bibinfo{person}{Yun Fu}, {and} \bibinfo{person}{Humphrey Shi}.} \bibinfo{year}{2022}\natexlab{}.
\newblock \showarticletitle{Towards layer-wise image vectorization}. In \bibinfo{booktitle}{\emph{Proceedings of the IEEE/CVF Conference on Computer Vision and Pattern Recognition}}. \bibinfo{pages}{16314--16323}.
\newblock


\bibitem[Myronenko and Song(2010)]%
        {myronenko2010point}
\bibfield{author}{\bibinfo{person}{Andriy Myronenko} {and} \bibinfo{person}{Xubo Song}.} \bibinfo{year}{2010}\natexlab{}.
\newblock \showarticletitle{Point set registration: Coherent point drift}.
\newblock \bibinfo{journal}{\emph{IEEE transactions on pattern analysis and machine intelligence}} \bibinfo{volume}{32}, \bibinfo{number}{12} (\bibinfo{year}{2010}), \bibinfo{pages}{2262--2275}.
\newblock


\bibitem[Nichol et~al\mbox{.}(2021)]%
        {nichol2021glide}
\bibfield{author}{\bibinfo{person}{Alex Nichol}, \bibinfo{person}{Prafulla Dhariwal}, \bibinfo{person}{Aditya Ramesh}, \bibinfo{person}{Pranav Shyam}, \bibinfo{person}{Pamela Mishkin}, \bibinfo{person}{Bob McGrew}, \bibinfo{person}{Ilya Sutskever}, {and} \bibinfo{person}{Mark Chen}.} \bibinfo{year}{2021}\natexlab{}.
\newblock \showarticletitle{Glide: Towards photorealistic image generation and editing with text-guided diffusion models}.
\newblock \bibinfo{journal}{\emph{arXiv preprint arXiv:2112.10741}} (\bibinfo{year}{2021}).
\newblock


\bibitem[Qin et~al\mbox{.}(2020)]%
        {qin2020u2}
\bibfield{author}{\bibinfo{person}{Xuebin Qin}, \bibinfo{person}{Zichen Zhang}, \bibinfo{person}{Chenyang Huang}, \bibinfo{person}{Masood Dehghan}, \bibinfo{person}{Osmar~R Zaiane}, {and} \bibinfo{person}{Martin Jagersand}.} \bibinfo{year}{2020}\natexlab{}.
\newblock \showarticletitle{U2-Net: Going deeper with nested U-structure for salient object detection}.
\newblock \bibinfo{journal}{\emph{Pattern recognition}}  \bibinfo{volume}{106} (\bibinfo{year}{2020}), \bibinfo{pages}{107404}.
\newblock


\bibitem[Quint(2003)]%
        {quint2003scalable}
\bibfield{author}{\bibinfo{person}{Antoine Quint}.} \bibinfo{year}{2003}\natexlab{}.
\newblock \showarticletitle{Scalable vector graphics}.
\newblock \bibinfo{journal}{\emph{IEEE MultiMedia}} \bibinfo{volume}{10}, \bibinfo{number}{3} (\bibinfo{year}{2003}), \bibinfo{pages}{99--102}.
\newblock


\bibitem[Radford et~al\mbox{.}(2021)]%
        {radford2021learning}
\bibfield{author}{\bibinfo{person}{Alec Radford}, \bibinfo{person}{Jong~Wook Kim}, \bibinfo{person}{Chris Hallacy}, \bibinfo{person}{Aditya Ramesh}, \bibinfo{person}{Gabriel Goh}, \bibinfo{person}{Sandhini Agarwal}, \bibinfo{person}{Girish Sastry}, \bibinfo{person}{Amanda Askell}, \bibinfo{person}{Pamela Mishkin}, \bibinfo{person}{Jack Clark}, {et~al\mbox{.}}} \bibinfo{year}{2021}\natexlab{}.
\newblock \showarticletitle{Learning transferable visual models from natural language supervision}. In \bibinfo{booktitle}{\emph{International conference on machine learning}}. PMLR, \bibinfo{pages}{8748--8763}.
\newblock


\bibitem[Ramesh et~al\mbox{.}(2021)]%
        {ramesh2021zero}
\bibfield{author}{\bibinfo{person}{Aditya Ramesh}, \bibinfo{person}{Mikhail Pavlov}, \bibinfo{person}{Gabriel Goh}, \bibinfo{person}{Scott Gray}, \bibinfo{person}{Chelsea Voss}, \bibinfo{person}{Alec Radford}, \bibinfo{person}{Mark Chen}, {and} \bibinfo{person}{Ilya Sutskever}.} \bibinfo{year}{2021}\natexlab{}.
\newblock \showarticletitle{Zero-shot text-to-image generation}. In \bibinfo{booktitle}{\emph{International Conference on Machine Learning}}. PMLR, \bibinfo{pages}{8821--8831}.
\newblock


\bibitem[Reddy et~al\mbox{.}(2021)]%
        {reddy2021im2vec}
\bibfield{author}{\bibinfo{person}{Pradyumna Reddy}, \bibinfo{person}{Michael Gharbi}, \bibinfo{person}{Michal Lukac}, {and} \bibinfo{person}{Niloy~J Mitra}.} \bibinfo{year}{2021}\natexlab{}.
\newblock \showarticletitle{Im2vec: Synthesizing vector graphics without vector supervision}. In \bibinfo{booktitle}{\emph{Proceedings of the IEEE/CVF Conference on Computer Vision and Pattern Recognition}}. \bibinfo{pages}{7342--7351}.
\newblock


\bibitem[Reed et~al\mbox{.}(2016)]%
        {reed2016generative}
\bibfield{author}{\bibinfo{person}{Scott Reed}, \bibinfo{person}{Zeynep Akata}, \bibinfo{person}{Xinchen Yan}, \bibinfo{person}{Lajanugen Logeswaran}, \bibinfo{person}{Bernt Schiele}, {and} \bibinfo{person}{Honglak Lee}.} \bibinfo{year}{2016}\natexlab{}.
\newblock \showarticletitle{Generative adversarial text to image synthesis}. In \bibinfo{booktitle}{\emph{International conference on machine learning}}. PMLR, \bibinfo{pages}{1060--1069}.
\newblock


\bibitem[Ribeiro et~al\mbox{.}(2020)]%
        {ribeiro2020sketchformer}
\bibfield{author}{\bibinfo{person}{Leo Sampaio~Ferraz Ribeiro}, \bibinfo{person}{Tu Bui}, \bibinfo{person}{John Collomosse}, {and} \bibinfo{person}{Moacir Ponti}.} \bibinfo{year}{2020}\natexlab{}.
\newblock \showarticletitle{Sketchformer: Transformer-based representation for sketched structure}. In \bibinfo{booktitle}{\emph{Proceedings of the IEEE/CVF conference on computer vision and pattern recognition}}. \bibinfo{pages}{14153--14162}.
\newblock


\bibitem[Rombach et~al\mbox{.}(2022)]%
        {rombach2022high}
\bibfield{author}{\bibinfo{person}{Robin Rombach}, \bibinfo{person}{Andreas Blattmann}, \bibinfo{person}{Dominik Lorenz}, \bibinfo{person}{Patrick Esser}, {and} \bibinfo{person}{Bj{\"o}rn Ommer}.} \bibinfo{year}{2022}\natexlab{}.
\newblock \showarticletitle{High-resolution image synthesis with latent diffusion models}. In \bibinfo{booktitle}{\emph{Proceedings of the IEEE/CVF Conference on Computer Vision and Pattern Recognition}}. \bibinfo{pages}{10684--10695}.
\newblock


\bibitem[Ruiz et~al\mbox{.}(2022)]%
        {ruiz2022dreambooth}
\bibfield{author}{\bibinfo{person}{Nataniel Ruiz}, \bibinfo{person}{Yuanzhen Li}, \bibinfo{person}{Varun Jampani}, \bibinfo{person}{Yael Pritch}, \bibinfo{person}{Michael Rubinstein}, {and} \bibinfo{person}{Kfir Aberman}.} \bibinfo{year}{2022}\natexlab{}.
\newblock \showarticletitle{Dreambooth: Fine tuning text-to-image diffusion models for subject-driven generation}.
\newblock \bibinfo{journal}{\emph{arXiv preprint arXiv:2208.12242}} (\bibinfo{year}{2022}).
\newblock


\bibitem[Schaldenbrand et~al\mbox{.}(2022)]%
        {schaldenbrand2022styleclipdraw}
\bibfield{author}{\bibinfo{person}{Peter Schaldenbrand}, \bibinfo{person}{Zhixuan Liu}, {and} \bibinfo{person}{Jean Oh}.} \bibinfo{year}{2022}\natexlab{}.
\newblock \showarticletitle{Styleclipdraw: Coupling content and style in text-to-drawing translation}.
\newblock \bibinfo{journal}{\emph{arXiv preprint arXiv:2202.12362}} (\bibinfo{year}{2022}).
\newblock


\bibitem[Selinger(2003)]%
        {selinger2003potrace}
\bibfield{author}{\bibinfo{person}{Peter Selinger}.} \bibinfo{year}{2003}\natexlab{}.
\newblock \bibinfo{title}{Potrace: a polygon-based tracing algorithm}.
\newblock
\newblock


\bibitem[Song et~al\mbox{.}(2022)]%
        {song2022clipvg}
\bibfield{author}{\bibinfo{person}{Yiren Song}, \bibinfo{person}{Xning Shao}, \bibinfo{person}{Kang Chen}, \bibinfo{person}{Weidong Zhang}, \bibinfo{person}{Minzhe Li}, {and} \bibinfo{person}{Zhongliang Jing}.} \bibinfo{year}{2022}\natexlab{}.
\newblock \showarticletitle{CLIPVG: Text-Guided Image Manipulation Using Differentiable Vector Graphics}.
\newblock \bibinfo{journal}{\emph{arXiv preprint arXiv:2212.02122}} (\bibinfo{year}{2022}).
\newblock


\bibitem[Tian and Ha(2022)]%
        {tian2022modern}
\bibfield{author}{\bibinfo{person}{Yingtao Tian} {and} \bibinfo{person}{David Ha}.} \bibinfo{year}{2022}\natexlab{}.
\newblock \showarticletitle{Modern evolution strategies for creativity: Fitting concrete images and abstract concepts}. In \bibinfo{booktitle}{\emph{Artificial Intelligence in Music, Sound, Art and Design: 11th International Conference, EvoMUSART 2022, Held as Part of EvoStar 2022, Madrid, Spain, April 20--22, 2022, Proceedings}}. Springer, \bibinfo{pages}{275--291}.
\newblock


\bibitem[Vinker et~al\mbox{.}(2022)]%
        {vinker2022clipasso}
\bibfield{author}{\bibinfo{person}{Yael Vinker}, \bibinfo{person}{Ehsan Pajouheshgar}, \bibinfo{person}{Jessica~Y Bo}, \bibinfo{person}{Roman~Christian Bachmann}, \bibinfo{person}{Amit~Haim Bermano}, \bibinfo{person}{Daniel Cohen-Or}, \bibinfo{person}{Amir Zamir}, {and} \bibinfo{person}{Ariel Shamir}.} \bibinfo{year}{2022}\natexlab{}.
\newblock \showarticletitle{Clipasso: Semantically-aware object sketching}.
\newblock \bibinfo{journal}{\emph{ACM Transactions on Graphics (TOG)}} \bibinfo{volume}{41}, \bibinfo{number}{4} (\bibinfo{year}{2022}), \bibinfo{pages}{1--11}.
\newblock


\bibitem[Wang and Mahadevan(2008)]%
        {wang2008manifold}
\bibfield{author}{\bibinfo{person}{Chang Wang} {and} \bibinfo{person}{Sridhar Mahadevan}.} \bibinfo{year}{2008}\natexlab{}.
\newblock \showarticletitle{Manifold alignment using procrustes analysis}. In \bibinfo{booktitle}{\emph{Proceedings of the 25th international conference on Machine learning}}. \bibinfo{pages}{1120--1127}.
\newblock


\bibitem[Yang et~al\mbox{.}(2023)]%
        {yang2023subpixel}
\bibfield{author}{\bibinfo{person}{Jinfan Yang}, \bibinfo{person}{Nicholas Vining}, \bibinfo{person}{Shakiba Kheradmand}, \bibinfo{person}{Nathan Carr}, \bibinfo{person}{Leonid Sigal}, {and} \bibinfo{person}{Alla Sheffer}.} \bibinfo{year}{2023}\natexlab{}.
\newblock \showarticletitle{Subpixel Deblurring of Anti-Aliased Raster Clip-Art}. In \bibinfo{booktitle}{\emph{Computer Graphics Forum}}, Vol.~\bibinfo{volume}{42}. Wiley Online Library, \bibinfo{pages}{61--76}.
\newblock


\bibitem[Yang et~al\mbox{.}(2015)]%
        {yang2015effective}
\bibfield{author}{\bibinfo{person}{Ming Yang}, \bibinfo{person}{Hongyang Chao}, \bibinfo{person}{Chi Zhang}, \bibinfo{person}{Jun Guo}, \bibinfo{person}{Lu Yuan}, {and} \bibinfo{person}{Jian Sun}.} \bibinfo{year}{2015}\natexlab{}.
\newblock \showarticletitle{Effective clipart image vectorization through direct optimization of bezigons}.
\newblock \bibinfo{journal}{\emph{IEEE transactions on visualization and computer graphics}} \bibinfo{volume}{22}, \bibinfo{number}{2} (\bibinfo{year}{2015}), \bibinfo{pages}{1063--1075}.
\newblock


\end{thebibliography}

\begin{figure*}[tbp]
  \centering
  \includegraphics[width=1.0\linewidth]{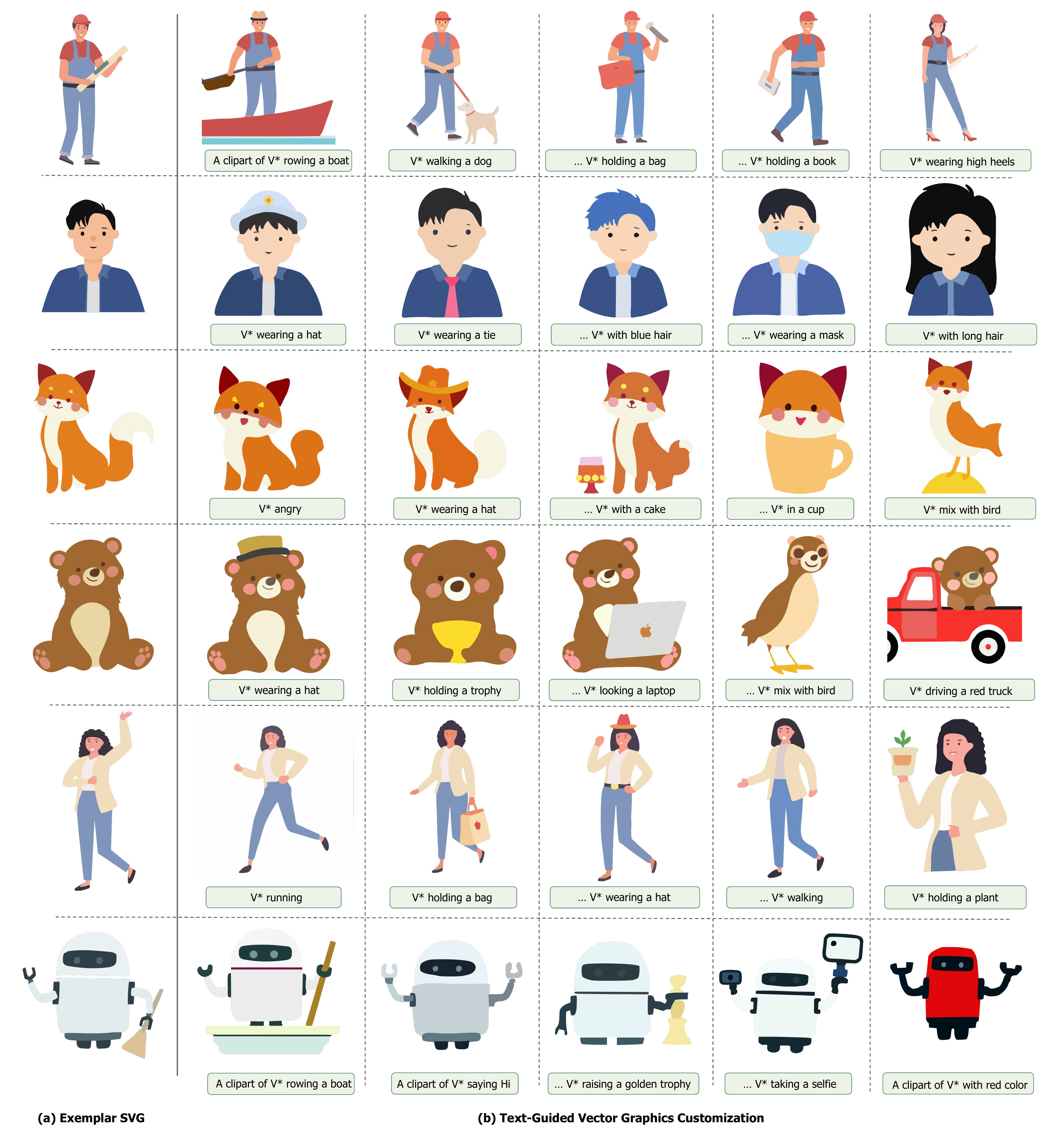}
  \caption{\label{fig:more_results} More results of our text-guided vector graphics customization. Exemplar SVGs: the $1^{st}$ and $5^{th}$ rows are from Freepik creator \copyright{Alexdndz}; the $2^{nd}$ row is from envatoelements creator \copyright{Telllu}; the $3^{rd}$ and $4^{th}$ rows are from \copyright{Freepik}; the $6^{th}$ row is from Iconduck creator \copyright{Pavan Kondapuram}.}
\end{figure*}

\end{document}